\documentclass[smallextended]{svjour3}
\usepackage{graphicx}
\usepackage{fancybox}
\usepackage{url}
\usepackage[utf8]{inputenc}
\usepackage{xspace}
\usepackage{amsmath}
\usepackage{amssymb}
\usepackage{hyperref}
\usepackage{booktabs}

\hypersetup{
  bookmarks=true,
  pdftitle={Premise Selection for Mathematics by Corpus Analysis and Kernel Methods},
  pdfauthor={Jesse Alama, Tom Heskes, Daniel Kuehlwein, Evgeni Tsivtsivadze, Josef Urban},
  pdfkeywords={automated theorem proving, interactive theorem proving, machine learning, proof analysis, formal mathematics, mizar},
  colorlinks=true,
  linkcolor=black,
  citecolor=black,
}
\usepackage{listings}
\def\systemname#1{\textsf{#1}\xspace}

\newcommand{\coloneqq}[0]{\mathrel{\mathop:}=}
\newcommand{\Vampire}[0]{\systemname{Vampire}}
\newcommand{\MaLARea}[0]{\systemname{MaLARea}}
\newcommand{\MOR}[0]{\systemname{MOR}}
\newcommand{\SNoW}[0]{\systemname{SNoW}}
\newcommand{\mizar}{\systemname{Mizar}}

\newcommand{\MML}{\text{{\sc mml}}}

\newcommand{\MaLeCoP}{\systemname{MaLeCoP}}

\newcommand{\isabelle}{\systemname{Isabelle}}
\newcommand{\coq}{\systemname{Coq}}
\newcommand{\hol}{\systemname{HOL}}

\newcommand{\usedPremises}[1]{usedPremises(#1)}
\newcommand{\advisedPremises}[2]{advisedPremises({#1},{#2})}
\newcommand{\recall}[2]{recall({#1},{#2})}

\DeclareMathOperator*{\argmin}{arg\,min}

\usepackage{xcolor}

\title{Premise Selection for Mathematics by\\ Corpus Analysis and Kernel Methods}
\author{Jesse Alama$^{*}$
\and Tom~Heskes$^{**}$
\and Daniel K\"{u}hlwein$^{**}$
\and Evgeni Tsivtsivadze$^{**}$
\and Josef Urban$^{**}$}
\institute{
$^{*}$ \;Center for Artificial Intelligence, New University of  Lisbon. Funded by the FCT
    project ``Dialogical Foundations of Semantics'' (DiFoS) in the ESF
    EuroCoRes programme LogICCC (FCT LogICCC/0001/2007).  Research for this paper was partially done while a visiting
fellow at the Isaac Newton Institute for the Mathematical Sciences in
the programme `Semantics \& Syntax'.\\
$^{**}$ \;Intelligent Systems, Institute for Computing and Information Sciences, Radboud University  Nijmegen. Funded by the NWO projects ``Learning2Reason'' and ``MathWiki''.}
\authorrunning{Alama, Heskes, K\"{u}hlwein, Tsivtsivadze, and Urban}

\begin{document}

\maketitle

\begin{abstract}
  Smart premise selection is essential when using automated reasoning
  as a tool for large-theory formal proof
  development. 
  A good method
  for premise selection in complex mathematical libraries is the
  application of machine learning to large corpora of proofs.

  \indent This work develops learning-based premise selection in two
  ways.  First, a newly available minimal dependency analysis of
  existing high-level formal mathematical proofs is used to build a
  large knowledge base of proof dependencies, providing precise
  data for ATP-based re-verification and for training premise
  selection algorithms.  Second, a new machine learning algorithm for
  premise selection based on kernel methods is proposed and
  implemented.  To evaluate the impact of both techniques, a 
  benchmark consisting of 2078 large-theory mathematical problems is
  constructed, extending the older MPTP Challenge benchmark.
  The combined effect of the techniques results in a 50\%
  improvement on the benchmark over the Vampire/SInE state-of-the-art system
  for automated reasoning in large theories.

\end{abstract}

\section{Introduction}

\label{Introduction}
In this paper we significantly improve theorem proving in large formal
mathematical libraries by using a two-phase approach combining precise proof
analysis with machine learning of premise selection.

The first phase makes the first practical use of the
newly available \emph{minimal dependency analysis} of the proofs in the large
\mizar{} Mathematical Library~(\MML)\footnote{\url{http://www.mizar.org}}. This analysis allows us to
construct precise problems for ATP-based re-verification of the \mizar
proofs.  More importantly, the precise dependency data can be used as
a large repository of previous problem-solving knowledge from which
premise selection can be efficiently automatically learned by machine learning algorithms.

In the second phase, a complementary improvement is achieved by using a new
kernel-based machine learning algorithm, which 
outperforms existing methods for premise selection.
This means
that based on the large number of previously solved mathematical
problems, we can more accurately estimate which premises will be
useful for proving a new conjecture.

Such learned knowledge considerably helps automated proving of
new formally expressed mathematical problems by recommending the most
relevant previous theorems and definitions from the very large
existing libraries, and thus shielding the existing ATP methods from considering
thousands of irrelevant axioms. The
better such symbiosis of formal mathematics and learning-assisted
automated reasoning gets, the better for both parties: improved
automated reasoning increases the efficiency of formal mathematicians,
and lowers the cost of producing formal mathematics. This in turn
leads to larger corpora of previously solved nontrivial problems from
which the learning-assisted ATP can extract additional problem-solving
knowledge covering larger and larger parts of mathematics.

The rest of the paper is organized as follows. Section~\ref{large}
describes recent developments in large-theory automated reasoning
and motivates our problem.
Section~\ref{Dependencies} summarizes the recent implementation of
precise dependency analysis over the large \MML, and its use for
ATP-based cross-verification and training premise selection.
Section~\ref{ML} describes the general machine learning approach to
premise selection and an efficient
kernel-based multi-output ranking algorithm for premise
selection. In Section~\ref{sec:Data} a new large-theory benchmark of 2078 related \MML{} problems
is defined, extending the older and smaller MPTP Challenge benchmark, and our techniques are evaluated
on this benchmark in Section~\ref{Experiments-and-Results}.
Section~\ref{Conclusion} concludes and discusses future work and
extensions.

\section{Automated Reasoning in Large Theories (ARLT)}
\label{large}

In recent years, large formal libraries of re-usable knowledge expressed in rich
formalisms have been built with interactive proof assistants, such as
\mizar{}~\cite{mizar-in-a-nutshell}, \isabelle~\cite{NipkowPW02}, \coq~\cite{BC04}, \hol~(\textsf{light})~\cite{HarrisonSA06}, and
others.  Formal approaches are also being used increasingly in
non-mathematical fields such as software and hardware verification and
common-sense reasoning about real-world knowledge.  Such trends lead
to growth of formal knowledge bases in these fields.

One important development is that a number of these formal knowledge
bases and core logics have been translated to first-order formats
suitable for ATPs~\cite{MengP08,Urban06,PeaseS07}, and first-order ATP is today 
routinely used for proof assistance in systems like Isabelle~\cite{PaulsonS07,BlanchetteBN11}, Mizar~\cite{UrbanS10,abs-1109-0616}, and HOL~\cite{Harrison96}.  These first-order translations give rise to large,
semantically rich corpora that present significant new challenges for
the field of automated reasoning.  The techniques developed so far for
ATP in large theories can be broadly divided into two categories:
\begin{enumerate}
\item Heuristic symbolic analysis of the formulas appearing in problems, and
\item Analysis of previous proofs.
\end{enumerate}
In the first category, the SInE preprocessor by K.~Hoder~\cite{HoderV11,UrbanHV10} has so far been the most successful.  SInE is
particularly strong in domains with many hierarchical definitions such
as those in common-sense ontologies.  In the second category, machine learning of premise
selection, as done e.g. by the \MaLARea~\cite{US+08} system, is an
effective method in hard mathematical domains, where the knowledge bases
contain proportionally many
more nontrivial lemmas and theorems than simple definitions,
and
previous verified proofs can be used for learning proof
guidance.

Automated reasoning in large mathematical corpora is an interesting
new field in several respects. Large theories permit data-driven
approaches~\cite{KMB} to constructing ATP algorithms; indeed, the
sheer size of such libraries actually necessitates such methods. It
turns out that purely deductive, brute-force search methods can be
improved significantly by heuristic and inductive\footnote{The word \emph{inductive} denotes here \emph{inductive reasoning}, as opposed to \emph{deductive reasoning}.} methods, thus
allowing experimental research into 
combinations~\cite{US+08} of inductive and deductive methods. Large-theory benchmarks like the MPTP
Challenge\footnote{\url{http://www.tptp.org/MPTPChallenge}}, and its extended version developed here in Section~\ref{sec:Data}, 
can serve for rigorous
evaluation of such novel Artificial Intelligence (AI) methods over thousands of real-world
mathematical problems.\footnote{We do not evaluate on the CASC LTB datasets,
because they are too small to allow machine learning techniques.
Our goal is to help
  mathematicians who work with and re-use large amounts of
  previously established complex proofs and theorems.
}
Apart from the
\textit{novel AI} aspect, and the obvious \textit{proof assistance}
aspect, automated reasoning over large formal mathematical corpora can
also become a new tool in the established field of \textit{reverse
  mathematics}~\cite{simpson-sosoa}. This line of research has been
already started, for example by Solovay's
analysis~\cite{solovay-fom-email} of the connection between Tarski's
axiom~\cite{tarski}
and the axiom of choice, and
by Alama's analysis of the Euler's polyhedron formula~\cite{alama-dissertation}, both conducted
over the \MML.

\section{Computing Minimal Dependencies in \mizar}
\label{Dependencies}

In the world of automated theorem proving, proofs contain essentially
all logical steps, even very small ones (such as the steps taken in a
resolution proof).  In the world of interactive theorem proving, one
of the goals is to allow the users to express themselves with minimal
verbosity.  Towards that end, interactive theorem proving (ITP) systems often come with
mechanisms for suppressing some steps of an argument.  By design, an
ITP can suppress logical and mathematical steps that might be
necessary for a complete analysis of what a particular proof depends
upon.  In this section we summarize a recently developed solution to
this problem for the 
\MML. The basis of the
solution is refactoring of the articles of the \MML{}
into one-item micro-articles, and computing their minimal dependencies
by a brute-force minimization algorithm.  For a more detailed
discussion of \mizar, see~\cite{mizar-first-30,mizar-in-a-nutshell};
for a more detailed discussion of refactoring and minimization
algorithms, see~\cite{alama-mamane-urban2011}.

As an example of how inferences in ITP-assisted formal
mathematical proofs can be suppressed, consider a theorem of the form
\[
\forall x:\tau [\psi\left ( g(x) \right )],
\]
where ``$x:\tau$'' means that the variable $x$ has type $\tau$,
and $g$ is a unary function symbol that accepts arguments of type
$\tau'$.  Suppose further that, prior to the assertion of this theorem,
it is proved that $\tau$ is a subtype of $\tau'$.  The well-formedness of the theorem depends on this subtyping
relationship.  Moreover, the proof of the theorem may not mention this
fact; the subtyping relationship between $\tau$ and $\tau'$ may very
well not be an outright theorem.  In such a situation, the fact
\[
\forall x (x : \tau \rightarrow x : \tau')
\]
is suppressed.  We can see that by not requiring the author of
a formal proof to supply such subtyping relationships, we permit him
to focus more on the heart of the matter of his proof, rather than
repeating the obvious.
But if we are interested in giving a complete answer
to the question of what a formalized proof depends upon, we must
expose suppressed facts and inferences. Having the complete answer is
important for a number of applications,
see~\cite{alama-mamane-urban2011} for examples. The particular
importance for the work described here is that when efficient
first-order ATPs are used to assist high-level formal proof assistants
like \mizar, the difference between the \emph{implicitly} used facts
and the \emph{explicitly} used facts disappears. The ATPs
need to explicitly know all
the facts that are necessary for finding the proofs.  (If we were to
omit the subtyping axiom, for example, an ATP might find that the problem is 
countersatisfiable.)

The first step in the computation of fine-grained dependencies in
\mizar{} is to break up each article in the \MML{} into a sequence of
\mizar{} texts, each consisting of a single top-level item (e.g.,
theorem, definition).
Each of these texts can---with suitable preprocessing---be regarded as a
complete, valid \mizar{} article in its own right. The decomposition
of a whole article from the \MML{} into such smaller articles
typically requires a number of nontrivial refactoring steps,
comparable, e.g., to automated splitting and re-factoring of large
programs written in programming languages with complicated syntactic
mechanisms. 

In \mizar{}, every article begins with a so-called \emph{environment}
specifying the background knowledge (theorems, notations, etc.) that
is used to verify the article.  The actual \mizar{} content that is
imported, given an environment, is, in general, a rather conservative
overestimate of the items that the article actually needs.  That is
why we apply a greedy minimization process to the environment to compute a minimal set of items that are
sufficient to verify each ``micro-article''.  This produces a minimal
set of dependencies\footnote{Precisely, the minimality means that removing any dependence will cause the verification to fail.} 
for each \mizar item, both \emph{syntactic} (e.g.,
notational macros), and \emph{semantic} (e.g., theorems, typings,
etc.). The drawback of this minimization process is that the
greedy approach to minimization\footnote{The basic greedy minimization proceeds by checking if an article still compiles after removing increasingly larger parts of the environment.} of certain kinds of dependencies can be time consuming.\footnote{This can be improved by heuristics for guessing the
needed dependencies, analogous to those used for ATP premise selection.} The advantage is that (unlike in any other proof
assistant) the computed set of dependencies is truly minimal (with respect to
the power of the proof checker), and does not include redundant
dependencies which are typically drawn in by overly powerful proof
checking algorithms (like congruence closure over sets of all available
equalities, etc.) when the dependency tracking is implemented
internally inside a proof assistant. The dependency minimization is particularly important for the ATP and premise-selection applications that are explained in this paper: a day more of routine computation of the minimal dependencies is a very good time investment if it can provide better guidance for the fast-growing search space explored by ATPs.
Another advantage of this approach is that it also provides
\emph{syntactic} dependencies, which are needed for real-world
recompilation of the particular item as written in the article. This
functionality is important for fast fine-grained recompilation in
formal wikis~\cite{AB+11}, however for semantic applications like ATP we are only
considering the truly semantic dependencies, i.e., those dependencies
that result in a formula when translated by the MPTP system~\cite{Urban06} to
first-order logic.

\begin{table}[htbp]
  \centering

  \caption{Effectiveness of fine-grained dependencies on the 33 MPTP2078 articles ordered from top to bottom by their order in the \MML{}.}
  \begin{tabular}{lrrrrr}
\toprule
    Article & Theorems  & Expl. Refs. & Uniq. Expl. Refs. & Fine
    Deps. & MPTP Deps.\\\midrule
 \texttt{xboole\_0} & 7 & 4 & 2.7 & 11.57 & 12.62 \\
 \texttt{xboole\_1} & 117 & 5.34 & 2.3 & 15.27 & 17.86 \\
 \texttt{enumset1} & 87 & 3.26 & 2.7 & 10.67 & 10.82 \\
 \texttt{zfmisc\_1} & 129 & 4.74 & 2.9 & 16.59 & 21.08 \\
 \texttt{subset\_1} & 43 & 4.62 & 2.4 & 22.30 & 28.15 \\
 \texttt{setfam\_1} & 48 & 6.56 & 2.4 & 25.62 & 37.93 \\
 \texttt{relat\_1} & 184 & 5.66 & 2.2 & 19.97 & 27.31 \\
 \texttt{funct\_1} & 107 & 7.69 & 3.4 & 22.94 & 42.05 \\
    \texttt{ordinal1} & 37 & 7.81 & 3.9 & 26 & 61.92 \\
  \texttt{wellord1} & 53 & 11.7 & 5.6 & 30.45 & 55.5 \\
    \texttt{relset\_1} & 32 & 4.71 & 2.6 & 27 & 55.59 \\
 \texttt{mcart\_1} & 92 & 5.71 & 2.8 & 21.25 & 29.77 \\
  \texttt{wellord2} & 24 & 14.2 & 6.9 & 36.41 & 75.2 \\
 \texttt{funct\_2} & 124 & 4.14 & 2.5 & 30.77 & 92.45 \\
 \texttt{finset\_1} & 15 & 8.66 & 3.9 & 23.93 & 72.22 \\
 \texttt{pre\_topc} & 36 & 6.47 & 3.2 & 35.58 & 53.51 \\
 \texttt{orders\_2} & 56 & 10.6 & 5.3 & 46.28 & 79.77 \\
    \texttt{lattices} & 27 & 5.59 & 3.1 & 43 & 72.96 \\
 \texttt{tops\_1} & 71 & 6.67 & 4.2 & 43.46 & 66.42 \\
 \texttt{tops\_2} & 65 & 7.36 & 4.0 & 37.41 & 103.4 \\
 \texttt{compts\_1} & 23 & 17.7 & 8.6 & 48.86 & 102.9 \\
  \texttt{connsp\_2} & 29 & 9.86 & 6.4 & 39.51 & 96.5 \\
 \texttt{filter\_1} & 61 & 14.6 & 5.3 & 52.27 & 122.8 \\
  \texttt{lattice3} & 55 & 8.6 & 4.2 & 47.07 & 92.25 \\
 \texttt{yellow\_0} & 70 & 6.75 & 3.2 & 26.55 & 46.12 \\
 \texttt{yellow\_1} & 28 & 9.03 & 5.1 & 53.17 & 128.3 \\
 \texttt{waybel\_0} & 76 & 9.34 & 4.1 & 35.03 & 82.34 \\
 \texttt{tmap\_1} & 141 & 8.78 & 4.5 & 47.04 & 140.0 \\
 \texttt{tex\_2} & 74 & 7.66 & 5.0 & 37.31 & 155.6 \\
 \texttt{yellow\_6} & 44 & 11.2 & 6.4 & 50.31 & 173.5 \\
 \texttt{waybel\_7} & 46 & 13.0 & 7.7 & 57.10 & 140.3 \\
    \texttt{waybel\_9} & 41 & 9 & 5.2 & 51.56 & 156.5 \\
 \texttt{yellow19} & 36 & 8.44 & 5.3 & 53.02 & 137.2 \\\bottomrule
\end{tabular}
{\small
  \begin{description}
  \item[Article:] \mizar{} Article relevant to the MPTP2078 benchmark.
  \item[Theorems:] Total number of theorems in the
    article.
  \item[Expl. Refs.:] Average number of (non-unique) explicit references
    (to theorems, definitional theorems, and schemes) per theorem in the
    article.
  \item[Uniq. Expl. Refs.:] Average number of unique explicit references per theorem.
  \item[Fine Deps.:]  Average number of all (both
    explicitly and implicitly) used items (explicitly referred to
    theorems, together with implicitly used items) per theorem as
    computed by dependency analysis.
  \item[MPTP Deps.:] Average number of items 
    per theorem as approximated by the MPTP fixpoint algorithm.
  \end{description}}
\label{fig:mptp-dep-data}
\end{table}

Table~\ref{fig:mptp-dep-data} provides a summary of the fine-grained
dependency data for the set of 33
\mizar{} articles coming from the MPTP2078 benchmark developed in Section~\ref{sec:Data}, and used for the experiments in Section~\ref{Experiments-and-Results}.
For each theorem in the sequence of the 33 \mizar{} articles (ordered from first to last by their order in the \MML{}) we show how many explicit dependencies are involved (on
average) in their proofs and how many implicit dependencies (on
average) it contains.    The table also shows how much
of an improvement the exact dependency calculation is, compared to a
simple safe fixed-point MPTP construction of an over-approximation of what is
truly used in the \MML{} proof.

\section{Premise Selection in Large Theories by Machine Learning}
\label{ML}
When reasoning over a large theory (like the \MML{}), thousands of premises are available.
In the presence of such large numbers of premises, the performance of most ATP systems degrades considerably~\cite{UrbanHV10}. Yet typically only a fraction of the available premises are actually needed to construct a proof.
Estimating which premises are likely to be useful for constructing a proof is our research problem:
\begin{definition}[Premise selection problem]\\
\emph{Given a large number of premises $\mathcal{P}$ and a new conjecture $c$, predict those premises from $\mathcal{P}$ that are likely to 
be useful for automatically constructing a proof of $c$.
}
\end{definition}

Knowledge of previous proofs and problem-solving techniques is used by mathematicians to guide their thinking about new problems.
The detailed \MML{} proof analysis described above provides a large computer-understandable corpus of dependencies of mathematical proofs.
In this section we present the machine learning setting and algorithms that are used to train premise selection on such corpora. Our goal is to begin emulating the training of human mathematicians.

When the translation \cite{Urban06} from \mizar to ATP formats is applied, the \mizar theorems and their proof dependencies (definitions, other theorems, etc.) 
translate to first-order formulas, used in the corresponding ATP problems as \emph{conjectures} and their \emph{premises} (axioms). For further presentation here we identify each \MML{} formula with its first-order translation.\footnote{\mizar is ``nearly first-order'', so the correspondence is ``nearly'' one-to-one, and in particular it is possible to construct the set of ATP premises from the exact \mizar dependencies for each \mizar theorem.}
We will work in the following setting, which is tailored to the \MML{}, but can easily be translated to other large datasets.
Let $\Gamma$ be the set of first order formulas that appear in the \MML{}.

\begin{definition}[Proof matrix]
  Using the fine-grained \mizar{} proof analysis - which says for
  each pair of formulas $p, c \in \Gamma$ 
whether $p$ is used
  in the \MML{} proof of $c$ - define the function $\mu : \Gamma \times
  \Gamma \rightarrow \{0,1\}$ by
\begin{equation*}
\mu(c,p) \coloneqq
\begin{cases}
1 & \text{if $p$ is used to prove $c$,}\\
0 &\text{otherwise.}
\end{cases}
\end{equation*}
\end{definition}
In other words, $\mu$ is the adjacency matrix of the graph of the
direct \MML{} proof dependencies. This proof matrix, together with
suitably chosen formula features, will be used for training machine
learning algorithms.

Note that in the \MML{}, there is always exactly one (typically a
textbook) proof of a particular theorem $c$, and hence exactly one set
of premises \linebreak $usedPremises(c) \coloneqq \left\{p \mid \mu(c,p) = 1 \right\}$ used in the proof
of $c$.  This corresponds to the mathematical textbook practice where
typically only one proof is given for a particular theorem. It is
however obvious that for example any expansion of the proof
dependencies can lead to an alternative proof.

In general, given a mathematical theory, there can be a variety of
more or less related alternative proofs of a particular theorem. This
variety however typically is not the \emph{explicit} textbook data on
which mathematicians study. Such variety is only formed (in different
measure) in their minds, after studying (training on) the textbook
proofs, which typically are chosen for some nice properties (simplicity, beauty,
clarity, educational value, etc.). Hence it is also plausible to use
the set of \MML{} proofs for training algorithms that attempt to
emulate human proof learning. Below we will often refer to the set of premises used in the (unique) \MML{} proof of a theorem $c$
as \emph{the} set of premises of $c$. This concept
is not to be read as the \emph{only} set of premises of $c$, but
rather as \emph{the particular set of premises that is used in human
  training}, and therefore is also likely to be useful in training
computers. It would not be difficult to relax this approach, if the
corpora from which we learn contained a number of good alternative
proofs. This is however so far not the case with the current \MML{},
on which we conduct these experiments.

Also note that although our training set consists of formal proofs, these
proofs have been authored by humans, and not found fully automatically
by ATPs.  But the evaluation conducted here (Section
\ref{Experiments-and-Results}) is done by running ATPs on the
recommended premises. It could be the case (depending on the ATP
implementation) that a fully automatically found proof would provide a
better training example than the human proof from the \MML{}. A major
obstacle for such training is however the relative weakness of
existing ATPs in finding more involved proofs of \MML{} theorems (see
\cite{UrbanHV10}), and thus their failure to provide the training
examples for a large part of \MML{}. Still, a comparison of the power
of training on \MML{} and ATP proofs could be interesting future work.

\begin{definition}[Feature matrix]
\label{features}
We characterize \MML{} formulas by the symbols and (sub)terms
appearing in them. We use de Bruijn indices 
for variables, and term
equality is then just string equality.
Let
$T \coloneqq \{ t_{1},\dots,t_{m} \}$ be a fixed enumeration of the
set of all symbols and (sub)terms that appear in all formulas from $\Gamma$.  We define
$\Phi : \Gamma \times \{1,\dots,m \} \rightarrow \{ 0,1 \}$ by
\begin{equation*}
\Phi(c,i) \coloneqq
\begin{cases}
1 & \text{if $t_i$ appears in $c$,}\\
0 &\text{otherwise.}
\end{cases}
\end{equation*}
This matrix gives rise to the \emph{feature function} $\boldsymbol{\varphi} : \Gamma \rightarrow \{0,1\}^m$ which for $c \in \Gamma$ 
is the vector $\boldsymbol{\varphi}^c$ with entries in $\{0,1\}$ satisfying
\[
\varphi^c_i = 1 \iff \Phi(c,i) = 1 .
\]
The \emph{expressed features of a formula} are denoted by the value of the function $e : \Gamma \rightarrow \mathcal{P}(T)$ that maps $c$ to $\{t_i \mid \Phi(c,i) = 1 \}$.
\end{definition}

Note that our choice of feature characterization is quite
arbitrary. We could try to use only symbols, or only (sub)terms, or
some totally different features. The better the features correspond to
the concepts that are relevant when choosing theorems
for solving a particular problem, the more successful the machine
learning of premise selection can be. As with the case of using 
alternative proofs for training, we just note that finding suitable
feature characterizations is a very interesting problem in this area,
and that our current choice seems to perform already quite reasonably
in the experiments. For the particular heuristic justification of
using formula (sub)terms, see~\cite{US+08}.

The premise selection problem can be treated as a ranking problem, or
as a classification problem.  In the ranking approach, we for a given
a conjecture $c$ rank the available premises by their predicted
usefulness for an automated proof of $c$, and use some number $n$ of
premises with the highest ranking (denoted here as $\advisedPremises{c}{n}$). In the classification approach, we
are looking for each premise $p \in \Gamma$ for a real-valued
\emph{classifier function} $C_{p}(\cdot): \Gamma \rightarrow
\mathbb{R}$ which, given a conjecture $c$, estimates how useful $p$ is
for proving $c$. In standard classification, a premise $p$ would then
be used if $C_{p}(c)$ is above certain threshold.  A common approach
to ranking is to use classification, and to combine the real-valued
classifiers~\cite{MLPprob2}: the premises for a conjecture $c$
are ranked by the values of $C_{p}(c)$, and we choose a certain number
of the best ones. This is the approach that we use in this paper.

Given a training corpus, machine learning algorithms can automatically \emph{learn} classifier functions.
The main difference between learning algorithms is the function space in which they search for the classifiers and the measure they use to evaluate how good a classifier is.  In our prior work on the applications of machine learning techniques to the premise selection problem~\cite{US+08} we used the \SNoW implementation~\cite{Carlson1999} of a multiclass naive Bayes learning method because of its efficiency.
In this work, we experiment with state-of-the-art \emph{kernel-based learning methods} for premise selection. We present both methods and show the benefits of using kernels.

\subsection{A Naive Bayes Classifier}
Naive Bayes is a statistical learning method based on Bayes's theorem about conditional probabilities\footnote{In its simplest form, Bayes's theorem asserts for a probability function $P$ and random variables $X$ and $Y$ that \[ P(X | Y) = \frac{P(Y|X)P(X)}{P(Y)}, \] where $P(X|Y)$ is understood as the conditional probability of $X$ given $Y$.} with a strong (read: naive) independence assumptions.
In the naive Bayes setting, the value $C_p(c)$ of the classifier function of a premise $p$ at a conjecture $c$ is the probability that $\mu(c,p) = 1$ given the expressed features $e(c)$.

To understand the difference between the naive Bayes and the kernel-based learning algorithm we need to take a closer look at the naive Bayes classifier.
Let $\theta$ denote the statement that $\mu(c,p) = 1$ and for each feature $t_i \in T$ let $\bar{t}_i$ denote that $\Phi(c,i) = 1$. Furthermore, let 
$e(c) = \{ s_1,\dots,s_l\} \subseteq T$ be the expressed features of $c$ (with corresponding $\bar{s}_1,\dots,\bar{s}_l$). We have

\begin{align}
P(\theta \mid \bar{s}_1,\dots,\bar{s}_l) &\varpropto \ln \frac{P(\theta \mid \bar{s}_1,\dots,\bar{s}_l)}{P(\neg \theta \mid \bar{s}_1,\dots,\bar{s}_l)} \\
&= \ln \frac{P(\bar{s}_1,\dots,\bar{s}_l \mid \theta) P(\theta)}{P(\bar{s}_1,\dots,\bar{s}_l \mid \neg \theta) P(\neg\theta)} \\
&= \ln \frac{P(\bar{s}_1,\dots,\bar{s}_l \mid \theta)}{P(\bar{s}_1,\dots,\bar{s}_l \mid \neg \theta)} + \ln\frac{P(\theta)}{P(\neg\theta)} \\
&= \ln \prod_{i = 1}^l \frac{P(\bar{s}_i \mid \theta)}{P(\bar{s}_i \mid \neg \theta)} + \ln \frac{P(\theta)}{P(\neg\theta)} \; \text{by independence}\\
&= \sum_{i = 1}^m \varphi^c_i \ln\left( \frac{P(\bar{t}_i \mid \theta)}{P(\bar{t}_i \mid \neg \theta)}\right) + \ln \frac{P(\theta)}{P(\neg\theta)} \\
&= \mathbf{w}^T \boldsymbol{\varphi}^c + \ln \frac{P(\theta)}{P(\neg\theta)} 
\end{align}

where 
\begin{equation}
\label{weights}
w_i \coloneqq \ln\left( \frac{P(\bar{t}_i \mid \theta)}{P(\bar{t}_i)\mid \neg \theta}\right)
\end{equation}
Line $(6)$ shows that the naive-Bayes classifier is ``essentially'' (after the monotonic transformation) a linear function of the features of the conjecture. The feature weights $\mathbf{w}$ are here computed using formula~(\ref{weights}).

\subsection{Kernel-based Learning}
We saw that the naive Bayes algorithm gives rise to a linear classifier. This leads to several questions: `Are there better parameters?' and `Can one get better performance with non-linear functions?'. Kernel-based learning provides a framework for investigating such questions.
In this subsection we give a simplified, brief description of kernel-based learning that is tailored to our present problem; for further information, see~\cite{aronszajn50reproducing,LearningKernels,KMB}.

\subsubsection{Are there better parameters?}
To answer this question we must first define what `better' means. Using the number of problems solved as measure is not feasible because we cannot practically run an ATP for every possible parameter combination. Instead, we measure how \emph{good} a classifier approximates our training data. We would like to have that
\[
\forall x \in \Gamma : C_p(x) = \mu(x,p).
\]
However, this will almost never be the case. To compare how well a classifier approximates the data, we use loss functions and the notion of expected loss that they provide, which we now define.

\begin{definition}[Loss function and Expected Loss]
A \emph{loss function} is any function $l : \mathbb{R} \times \mathbb{R} \rightarrow \mathbb{R}^+$.

Given a loss function $l$ we can then define the \emph{expected loss} $E(\cdot)$ of a classifier $C_p$ as
$$ E(C_p) = \sum_{x \in \Gamma} l(C_p(x),\mu(x,p))$$
\end{definition}
One might add additional properties such as $l(x,x) = 0$, but this is not necessary. Typical examples of a loss function $l(x,y)$ are the square loss $(y-x)^2$ or the $0$-$1$ loss defined by $I(x = y)$.

We can compare two different classifiers via their expected loss. If the expected loss of classifier $C_p$  is less than the expected loss of a classifier $C_q$ then $C_p$ is the better classifier. 
It should be noted that a lower expected loss on a particular training set 
(like the \MML{} proofs) need not necessarily lead to more solved problems by an ATP. 
One could imagine that the training set contains proofs that are very different from the way 
a particular ATP would proceed most easily. Also, what happens if the classifier is not able
to predict all \MML{} premises, but just a large part of them? These are questions about alternative
proofs, and about the robustness of the ATP and prediction methods. An experimental answer is 
provided in Section~\ref{together}.

\subsubsection{Nonlinear Classifiers}
It seems straightforward that more complex functions would lead to a lower expected loss and are hence desirable. However, parameter optimization becomes tedious once we leave the linear case. Kernels provide a way to use the machinery of linear optimization on non-linear functions.

\begin{definition}[Kernel]
 A \emph{kernel} is is a function $k : \Gamma \times \Gamma \rightarrow \mathbb{R}$ satisfying
$$k(x,y) = \langle \phi(x),\phi(y) \rangle$$
where $\phi : \Gamma \rightarrow F$ is a mapping from $\Gamma$ to an inner product space $F$ with inner product $\langle \cdot, \cdot \rangle$.
A kernel can be understood as a \emph{similarity measure} between two entities.
\end{definition}

\begin{example}\label{ex:one}
A simple kernel for our setting is the linear kernel:
$$k_{\text{lin}}(x,y) \coloneqq \langle \boldsymbol{\varphi}^x,\boldsymbol{\varphi}^y \rangle$$
with $\langle \cdot,\cdot \rangle$ being the normal dot product in $\mathbb{R}^m$.
Here, $\boldsymbol{\varphi}^f$ denotes the features of a formula $f$ (see definition \ref{features}), and the inner product space $F$ is $\mathbb{R}^m$.
A nontrivial example is the Gaussian kernel with parameter $\sigma$:
\[
k_{\text{gauss}}(x,y) \coloneqq \exp\left(-\frac{\langle \boldsymbol{\varphi}^x,\boldsymbol{\varphi}^x \rangle -  2 \langle \boldsymbol{\varphi}^x,\boldsymbol{\varphi}^y \rangle + \langle \boldsymbol{\varphi}^y,\boldsymbol{\varphi}^y \rangle}{\sigma^{2}}\right)
\]
\end{example}

We can now define our kernel function space in which we will search for classification functions.

\begin{definition}[Kernel Function Space]
Given a kernel $k$, we define
\[
\mathcal{F}_k \coloneqq \left\{ f \in \mathbb{R}^\Gamma \mid f(x) = \sum_{v \in \Gamma} \alpha_v k(x,v), \alpha_v \in \mathbb{R}, \Arrowvert f \Arrowvert < \infty \right\}.
\]
as our \emph{kernel function space}, where
\[
\left \Arrowvert \sum_{v \in \Gamma} \alpha_v k(x,v) \right \Arrowvert = \sum_{u,v \in \Gamma} \alpha_u \alpha_v k(u,v)
\]
Essentially, every function in $\mathcal{F}_k$ compares the input $x$ with formulas in $\Gamma$ using the kernel, and the weights $\alpha$ determine how important each comparison is\footnote{A more general approach to kernel spaces is available; see \cite{Schoelkopf2001}.}.
\end{definition}

The kernel function space $\mathcal{F}_{k}$ naturally depends on the
kernel $k$. It can be shown that when we use $k_{\text{lin}}$,
$\mathcal{F}_{k_\text{lin}}$ consists of linear functions of the
\MML{} features $T$. In contrast, the Gaussian kernel
$k_{\text{gauss}}$ gives rise to a very nonlinear kernel function
space.

\subsubsection{Putting it all together}
Having defined loss functions, kernels and kernel function spaces we can now define how kernel-based learning algorithms learn classifier functions.
Given a kernel $k$ and a loss function $l$, recall that we measure how good a classifier $C_p$ is with the expected loss $E(C_p)$.
With all our definitions it seems reasonable to define $C_p$ as

\begin{equation}
\label{simplecp}
C_p \coloneqq \argmin_{f \in \mathcal{F}_k} E(f)
\end{equation}
However, this is not what a kernel based learning algorithm does. There are two reasons for this.
First, the minimum might not exist. Second, in particular when using complex kernel functions, such an approach might lead to overfitting: $C_p$ might perform very well on our training data, but bad on data that was not seen before.
To handle both problems, a \emph{regularization parameter} $\lambda > 0$ is introduced to penalize complex functions (assuming that high complexity implies a high norm).
This regularization parameter allows us to place a bound on possible solution which together with the fact that $\mathcal{F}_k$ is a Hilbert space ensures the existence of $C_p$.
Hence we define
\begin{equation}\label{regproblem}
C_p = \argmin_{f \in \mathcal{F}_k} E(f)+\lambda\Arrowvert f \Arrowvert^2
\end{equation}

Recall from the definition of $\mathcal{F}_k$ that $C_p$ has the form
\begin{equation}
\label{alpharep}
C_p(x)=\sum_{v \in \Gamma} \alpha_v k(x,v),
\end{equation}
with $\alpha_v \in\mathbb{R}$. Hence, for any fixed $\lambda$, we only need to compute the weights $\alpha_v$ for all $v \in \Gamma$ in order to find $C_p$.
In section \ref{MORSetup} we show how to solve this optimization problem in our setting.

\subsubsection{Naive Bayes vs Kernel-based Learning}

Kernel-based methods typically outperform the naive Bayes algorithm.  There are several reasons for this. Firstly and most importantly, while naive Bayes 
is essentially a linear classifier, kernel based methods can learn
non-linear dependencies when an appropriate non-linear (e.g. Gaussian)
kernel function is used. This advantage in
expressiveness usually leads to significantly better 
generalization\footnote{\emph{Generalization} is the ability of a machine learning
algorithm to perform accurately on new, unseen examples after training on a
finite data set.}
performance of the algorithm given properly estimated hyperparameters
(e.g., the kernel width for Gaussian functions). Secondly, kernel-based
methods are formulated within the regularization framework that
provides mechanism to control the errors on the training set and the
complexity ("expressiveness") of the prediction function. Such setting
prevents overfitting of the algorithm and leads to notably better
results compared to unregularized methods. Thirdly, some of the
kernel-based methods (depending on the loss function) can use 
very efficient procedures for hyperparameter estimation (e.g. fast
leave-one-out cross-validation~\cite{Rifkin2003}) and therefore result in
a close to optimal model for the classification/regression task. For such reasons
kernel-based methods are among the most successful algorithms applied to
various problems from bioinformatics to information retrieval to
computer vision~\cite{KMB}.
A general
advantage of naive Bayes over kernel-based algorithms is the
computational efficiency, particularly when taking into account
the fact that computing the kernel matrix is generally quadratic in the number of training
data points. However, recent advances in large scale learning have led to
extensions of various kernel-based methods such as SVMs, with sublinear
complexity, provably fast convergence rate, and the generalization
performance that cannot be matched by most of the methods in the
field~\cite{Shalev-ShwartzSSC11}.

\subsection{\MOR Experimental Setup}
\label{MORSetup}
For our experiments, we will now define a kernel-based multi-output ranking (\MOR) algorithm that is a relatively straightforward extension of our preference learning algorithm
presented in \cite{PL10}. \MOR{} is also based on the regularized least-squares algorithm presented in~\cite{Rifkin2003}.

Let $\Gamma = \{ x_1, \dots , x_n\}$. Then formula (\ref{alpharep}) becomes
$$ C_p(x)=\sum_{i = 1}^n \alpha_i k(x,x_i) $$
Using this and the square-loss $l(x,y)=(x-y)^2$ function, solving equation (\ref{regproblem}) is equivalent to finding weights $\alpha_i$ that minimize
\begin{equation}\label{regsimple}
\min_{\alpha_1,\hdots,\alpha_n} \left [ \sum_{i=1}^n \left ( \sum_{j=1}^{n} \alpha_j k(x_i,x_j) - \mu(x_i,p) \right )^{2} + \lambda \sum_{i,j=1}^n \alpha_i \alpha_j k(x_i,x_j) \right ]
\end{equation}

Recall that $C_p$ is the classifier for a single premise. Since we eventually want to rank all premises, we need to train a classifier for each premise.
So we need to find weights $\alpha_{i,p}$ for each premise $p$. This does seem to complicate the problem quite a bit.
However, we can use the fact that for each premise $p$, $C_p$ depends on the values of $k(x_i,x_j)$, where $1 \leq i,j \leq n$, to speed up the computation.
Instead of learning the classifiers $C_p$ for each premise separately, we learn all the weights $\alpha_{p,i}$ simultaneously.

To do this, we first need some definitions.
Let
\[
A = (\alpha_{i,p})_{i,p} \quad (1 \leq i \leq n, p \in \Gamma).
\]
$A$ is the matrix where each column contains the parameters of one premise classifier.
Define the kernel matrix $K$ and the label matrix $Y$ as
\[
\begin{array}{lcl}
K & \coloneqq & (k(x_i,x_j))_{i,j} \quad (1 \leq i,j \leq n)\\
Y & \coloneqq & (\mu(x_i,p))_{i,p} \quad (1 \leq i \leq n, p \in \Gamma).
\end{array}
\]
We can now rewrite (\ref{regsimple}) in matrix notation to state the problem for all premises:
\begin{equation}\label{objectiveinmatrixform}
\argmin_A \text{tr}\left( (Y-KA)^\textrm{T}(Y-KA) + \lambda A^\textrm{T}KA \right)
\end{equation}
where $\text{tr}(A)$ denotes the trace of the matrix $A$. Taking the derivative with respect to $A$ leads to:
\begin{eqnarray*}
& \frac{\partial}{\partial A} \text{tr} \left( (Y-KA)^\textrm{T} (Y-KA)+\lambda A^\textrm{T}KA \right)\\
= &-2K(Y-KA)
+2\lambda KA\\
= &-2K Y
+(2KK+2\lambda K)A
\end{eqnarray*}
To find the minimum, we set the derivative to zero and solve with respect to $A$. This leads to:
\begin{eqnarray}
A
&=&(KK+\lambda K)^{-1} K Y\\
&=&(K+\lambda I)^{-1} Y
\end{eqnarray}

In the experiments, we use the Gaussian kernel $k_{\text{gauss}}$ we defined in Example~\ref{ex:one}.
Ergo, if we fix the regularization parameter $\lambda$ and the kernel parameter $\sigma$ we can find the optimal weights through simple matrix computations.
Thus, to fully determine the classifiers, it remains to find good values for the parameters $\lambda$ and $\sigma$.
This is done, as is common with such parameter optimization for kernel methods,
by simple (logarithmically scaled) grid search and cross-validation on the training data using a $70/30$ split.

\section{Data: The MPTP2078 Benchmark}
\label{sec:Data}
The effects of using the minimized dependency data (both
for direct re-proving and for training premise selection),
and the effect of using our kernel-based \MOR algorithm are evaluated on a
newly created large-theory benchmark\footnote{\url{http://wiki.mizar.org/twiki/bin/view/Mizar/MpTP2078}} of 2078 related \MML{} problems,
which extends the older and smaller MPTP Challenge benchmark.

The original MPTP Challenge benchmark was created in 2006, with the
purpose of supporting the development of ARLT (automated reasoning for
large theories) techniques. It contains 252 related problems, leading
to the Mizar proof of one implication of the Bolzano-Weierstrass
theorem.
The challenge has two divisions: \emph{chainy} (harder) and
\emph{bushy} (easier). The motivation behind them is given below when
we describe their analogs in the MPTP2078 benchmark.

Both the ARLT techniques and the computing power (particularly
multi-core technology) have developed since 2006. Appropriately, we
define a larger benchmark with a larger numbers of problems and
premises, and making use of the more precise dependency knowledge. The
larger number of problems together with their dependencies more
faithfully mirror the setting that mathematicians are facing:
typically, they know a number of related theorems and their proofs
when solving a new problem.

The new MPTP2078 benchmark is created as follows: The 33 \mizar
articles from which problems were previously selected for constructing
the MPTP Challenge are used. We however use a new version of \mizar
and \MML{} allowing the precise dependency analysis, and use all
problems from these articles. This yields 2078 problems. As with the
MPTP Challenge benchmark, we create two groups (divisions) of problems.
\begin{itemize}
\item [{\bf Chainy:}] Versions of the 2078 problems containing all previous \MML{}
  contents as premises. This means that the conjecture is attacked
  with ``all existing knowledge'', without any premise selection. This
  is a common use case for proving new conjectures fully
  automatically, see also Section~\ref{Combining}. In the MPTP
  Challenge, the name \emph{chainy} has been introduced for this
  division, because the problems and dependencies are ordered into a
  chronological chain, emulating the growth of the library.
\item [{\bf Bushy:}] Versions of the 2078 problems with premises pruned using the new
  fine-grained dependency information. This use-case has been
  introduced in proof assistants by Harrison's \texttt{MESON\_TACTIC}~\cite{Harrison96},
  which takes an explicit list of premises from the large library
  selected by a knowledgeable user, and attempts to prove the
  conjecture just from these premises. We are interested in how
  powerful ATPs can get on \MML{} with such precise advice.
\end{itemize}
To evaluate the benefit of having precise minimal dependencies, we
additionally also produce in this work versions of the 2078 problems
with premises pruned by the old heuristic dependency-pruning method
used for constructing re-proving problems by the MPTP system. The MPTP
heuristic proceeds by taking all \emph{explicit} premises contained in
the original human-written \mizar proof. To get all the premises used
by \mizar \emph{implicitly}, the heuristic watches the problem's set
of symbols, and adds the implicitly used formulas (typically typing
formulas about the problem's symbols) in a fixpoint manner. The
heuristic attempts hard to guarantee completeness, however, minimality
is not achievable with such simple approach.

All three datasets contain the same conjectures. They only differ in the
number of redundant axioms. Note that the problems in the second and third dataset are
considerably smaller than the unpruned problems. The average number of
premises is 1976.5 for the unpruned (chainy) problems, 74 for the
heuristically-pruned problems (bushy-old) and 31.5 for the problems pruned using
fine-grained dependencies (bushy). Table~\ref{tab:Data} summarizes the datasets.

\begin{table}[htbp]
\centering
\caption{Average Number of Premises in the three Datasets}
\begin{tabular}{llr}
\toprule
Dataset & Premises used & Avg. number of premises \\ \midrule
Chainy & All previous & 1976.5\\
Bushy-Old & Heuristic dependencies & 74 \\
Bushy  & Minimized dependencies & 31.5  \\
\bottomrule
\end{tabular}
\label{tab:Data}
\end{table}

\section{Experiments and Results}
\label{Experiments-and-Results}

We use \Vampire{} 0.6~\cite{Vampire} as the ATP system for all
experiments conducted here. Adding other ATP systems is useful (see,
e.g., \cite{UrbanHV10} for recent evaluation), and there are
metasystems like \MaLARea which attempt to exploit the joint power of
different systems in an organized way. However, the focus of this work
is on premise selection, which has been shown to have similar effect
across the main state-of-the-art ATP systems. Another reason for using
the recent \Vampire{} is that in~\cite{UrbanHV10}, Vampire with the SInE
preprocessor was sufficiently tested and tuned on the \MML{} data,
providing a good baseline for comparing learning-based
premise-selection methods with robust state-of-the-art methods that
can run on any isolated large problem without any learning.
All
measurements are done on an Intel Xeon E5520 2.27GHz server with 8GB
RAM and 8MB CPU cache. Each problem is always assigned one CPU.

In Section~\ref{dep-evaluation} we evaluate the ATP performance when 
fine-grained dependencies (bushy problems) are used by comparing it to the
ATP performance on the old MPTP heuristic pruning (bushy-old problems), and to
the ATP performance on the large (chainy) versions of the MPTP2078
problems.  These results show that there is a lot to gain by
constructing good algorithms for premise selection.  In
Section~\ref{Combining} \SNoW's naive Bayes and the
\MOR machine learning algorithms are incrementally trained on the
fine-grained \MML{} dependency data, and their precision in predicting
the \MML{} premises on new problems are compared. This standard
machine-learning comparison is then in Section~\ref{together}
completed by running \Vampire{} on the premises predicted by the \MOR and \SNoW algorithms. This provides
information about the overall theorem-proving performance of the whole
dependency-minimization/learning/ATP stack. This performance is  compared to the performance of
\Vampire{}/SInE.

\subsection{Using the Fine-Grained Dependency Analysis for Re-proving}
\label{dep-evaluation}
The first experiment evaluates the effect of fine-grained dependencies
on re-proving \mizar theorems automatically. The results of
\Vampire{}/SInE run with 10s time limit\footnote{There are several reasons why we use low time limits. First, \Vampire{} performs reasonably with them in~\cite{UrbanHV10}. Second, low time limits are useful when conducting large-scale experiments and combining different strategies. Third, in typical ITP proof-advice scenarios~\cite{abs-1109-0616}, the preferable query response time is in (tens of) seconds. Fourth, 10 seconds in 2011 is much more than it was fifteen years ago, when the CASC competition started.}
on the datasets defined above are shown in
Table~\ref{tab1}.

\begin{table*}[htbp]
  \caption{Performance of \Vampire{} (10s time limit) on 2078 MPTP2078 benchmark with different axiom pruning.}
\centering
  \begin{tabular}{lrr}
\toprule
Pruning & Solved problems & Solved as percentage  \\\midrule
Chainy & 548 & 26.4 \\
Chainy (Vampire \verb+-d1+) & 556 & 26.8 \\
Bushy-old & 1023 & 49.2 \\
Bushy & 1105 & 53.2 \\
\bottomrule
  \end{tabular}
  \label{tab1}
\end{table*}

\Vampire{} (run in
the unmodified automated CASC mode) solves 548 of
the unpruned problems. If we use the \verb+-d1+
parameter\footnote{The \texttt{-d} parameter limits the depth of recursion for the SInE algorithm. In~\cite{UrbanHV10} running \Vampire with the \texttt{-d1} pruning parameter resulted in significant performance improvement on
  large \mizar problems.}, \Vampire{} solves 556 problems.  Things
change a lot with external premise pruning. \Vampire{} solves 1023 of the 2078 problems
when the old MPTP heuristic pruning (bushy-old) is applied. Using the pruning based on
the new fine-grained
analysis \Vampire{} solves 1105 problems, which is an 8\% improvement
over the heuristic pruning in the number of problems solved. Since the
heuristic pruning becomes more and more inaccurate as the \MML{}
grows (the ratio of MPTP Deps. to Fine Deps. in Table~\ref{fig:mptp-dep-data} has a growing trend from top to bottom), we can conjecture that this improvement will be even more
significant when considering the whole \MML. Also note that these
numbers point to the significant improvement potential that can be
gained by good premise selection: the performance on the pruned
dataset is doubled in comparison to the unpruned dataset. Again, this
ratio grows as \MML{} grows, and the number of premises approaches
100.000.\footnote{In the evaluation done in~\cite{UrbanHV10} on the whole \MML{} with \Vampire{}/SInE, this ratio is 39\% to 14\%.}

\subsection{Combining Fine-Grained Dependencies with Learning}
\label{Combining}
For the next experiment, we emulate the growth of the library (limited to the 2078 problems), by
considering all previous theorems and definitions when a new
conjecture is attempted.  This is a natural ``ATP advice over the
whole library'' scenario, in which the ATP problems however become very large,
containing thousands of the previously proved formulas. Premise
selection can therefore help significantly.

We use the fine-grained \MML{} dependencies extracted from previous
proofs\footnote{We do not evaluate the performance of learning on the
  approximate bushy-old dependencies here and in the next
  subsection. Table~\ref{fig:mptp-dep-data} and
  Section~\ref{dep-evaluation} already sufficiently show that these
  data are less precise than the fine-grained \MML{} dependencies.}
to train the premise-selection algorithms, use their advice on the
new problems, and compare the recall (and also the ATP performance in
the next subsection).  For each problem, the learning algorithms are
allowed to learn on the dependencies of all previous problems, which
corresponds to the situation in general mathematics when
mathematicians not only know many previous theorems, but also re-use
previous problem solving knowledge.
This approach requires us to do 2078
training steps as the problems and their proofs are added to the library
and the dataset grows. We compare the \MOR{} algorithm with \SNoW{}'s naive Bayes.

\begin{figure}[!htb]
\begin{center}
\includegraphics[width=12cm]{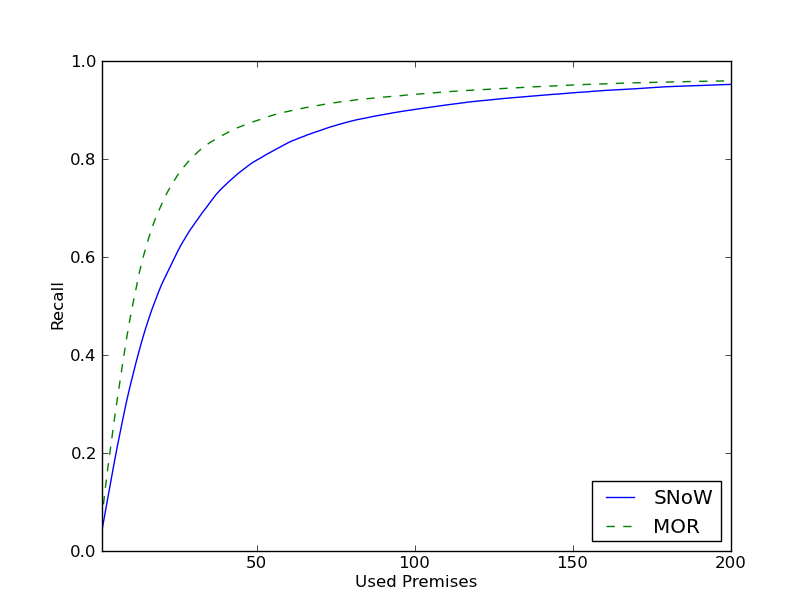}
\end{center}
\caption{Comparison of the \SNoW and \MOR average recall of premises used in the Mizar proofs. The x-axis shows the number of premises asked from \SNoW and \MOR, and the y-axis shows their relative overlap with the premises used in the original Mizar proof.}
\label{MPTPCover}
\end{figure}

Figure \ref{MPTPCover} shows the average recall of \SNoW and \MOR on this dataset.
The rankings obtained from the algorithms are compared
with the actual premises used in the \MML{} proof, by computing the size (ratio) of
the overlap for the increasing top segments of the ranked predicted
premises (the size of the segment is the x axis in
Figure~\ref{MPTPCover}). Formally, the recall $\recall{c}{n}$ for one conjecture $c$ when $n$ premises are advised is defined as:
\[
  \recall{c}{n}
  =
  \frac{\left | \usedPremises{c} \cap \advisedPremises{c}{n} \right |}%
       {\left | \usedPremises{c} \right |}
\]
It can be seen that the \MOR algorithm performs considerably better than \SNoW.
E.g., on average $88\%$ of the used premises are within the $50$ highest
\MOR-ranked premises, whereas when we consider the \SNoW ranking only around $80\%$ of the used premises are with the $50$ highest ranked premises.

Note that this kind of comparison is the standard
endpoint in machine learning applications like keyword-based document
retrieval, consumer choice prediction, etc. However, in a semantic
domain like ours, we can go further, and see how this improved
prediction performance helps the theorem proving process.
This is also interesting to see, because having for example only 90\% coverage of the original
\MML{} premises could be insufficient for constructing an ATP proof, unless the ATP
can invent alternative (sub-)proofs.\footnote{See~\cite{AlamaKU12} for an initial exploration of the phenomenon of alternative ATP proofs for \MML{} theorems.}
 This final evaluation is done
in the next section.

\subsection{Combining It All: ATP Supported by Learning from Fine Dependencies}
\label{together}
In the last experiment, we finally chain the whole ITP/Learning/ATP
stack together, and evaluate how the influence of the improved
premise selection reflects on performance of automated
theorem proving on new large-theory conjectures. Both the naive Bayes (\SNoW) and the new \MOR learning algorithms are evaluated.

\begin{figure}[!htb]
\begin{center}
\includegraphics[width=12cm]{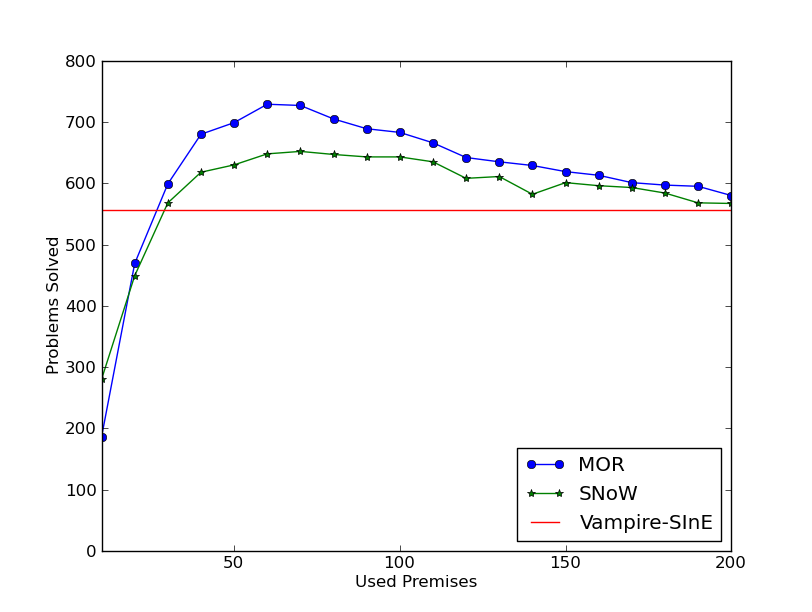}
\end{center}
\caption{Comparison of the number of solved problems by \SNoW and \MOR. The x-axis shows the number of premises given to the \Vampire, and the y-axis shows the number of problems solved within 5 seconds. The number of problems solved by \Vampire/SInE in 10 seconds is given as a baseline.}
\label{solvedFig}
\end{figure}

Figure \ref{solvedFig} shows the numbers of problems solved by \Vampire using different numbers of the top premises predicted by \SNoW and \MOR, and a 5 seconds time limit.
The maximum number of problems solved with \MOR is 729 with the top 60
advised premises.  \SNoW's maximum is 652 with the top 70
premises. The corresponding numbers for a 10 second time limit are 795
solved problems for \MOR{}-60, and 722 for \SNoW-70.
Table \ref{tab21} compares these data with the overall performance of \Vampire{} with a
10 second time limit run on problems with pruning done by SInE.
The
\SNoW{}-60 resp. \MOR{}-70 runs give a 32\% resp. 45\% improvement over the 548 problems \Vampire{}
solves in auto-mode, and a 30\% resp. 43\% improvement over the 556 problems
solved by \Vampire{} using the \verb+-d1+ option.

\begin{table}[htbp]
  \centering
  \caption{Comparison of \Vampire{} (10s time limit) performance on MPTP2078 with different premise selections.}
  \begin{tabular}{lrr}
\toprule
System & Solved Problems & Gain over \Vampire \\\midrule
\Vampire{}/SInE & 548 & 0 \% \\
\Vampire{}/SInE (\verb+-d1+)  & 556 & 1.5\% \\
\SNoW-70 & 722 & 31.8\% \\
\MOR-60 & 795 & 45.1\% \\
\bottomrule
  \end{tabular}
  \label{tab21}
\end{table}

Table \ref{tab2} additionally compares the performance of \Vampire{}/SInE 
with the performance of \SNoW{} and \MOR when 
computed for each of them as a union of the two 5s runs with the largest joint coverage. Those are obtained
by using the top 40 advised premises and the top 180 advised premises for \SNoW, and the top 40 and top 100 advised premises for \MOR.
These \SNoW resp. \MOR combined runs give a 44\% resp. 50\% improvement over the 548 problems \Vampire
solves in auto-mode, and a 42\% resp. 48\% improvement over the 556 problems
solved by \Vampire using the \verb+-d1+ option.

Note that \Vampire/SInE does strategy scheduling
  internally, and with different SInE parameters. Thus combining two
  different premise selection strategies by us is perfectly comparable
to the way \Vampire's automated mode is constructed and used. Also
note that combining the two different ways in which unadvised \Vampire/SInE was
run is not productive: the union of both unadvised runs is just
559 problems, which is only 3 more solved problems (generally in 20s)
than with running \Vampire/SInE with \verb+-d1+ for ten seconds.

\begin{table}[htbp]
  \centering
  \caption{10s performance of the two strategies with the largest joint coverage for \SNoW{} and \MOR.}
  \begin{tabular}{lrr}
\toprule
System & Solved Problems & Gain over \Vampire \\\midrule
\Vampire{}/SInE & 548 & 0 \% \\
\Vampire{}/SInE (\verb+-d1+)  & 556 & 1.5\% \\
\SNoW-40/180 & 788 & 43.7\% \\
\MOR-40/100 & 824 & 50.4\% \\
\bottomrule
  \end{tabular}
  \label{tab2}
\end{table}

\begin{figure}[!htb]
\begin{center}
\includegraphics[width=12cm]{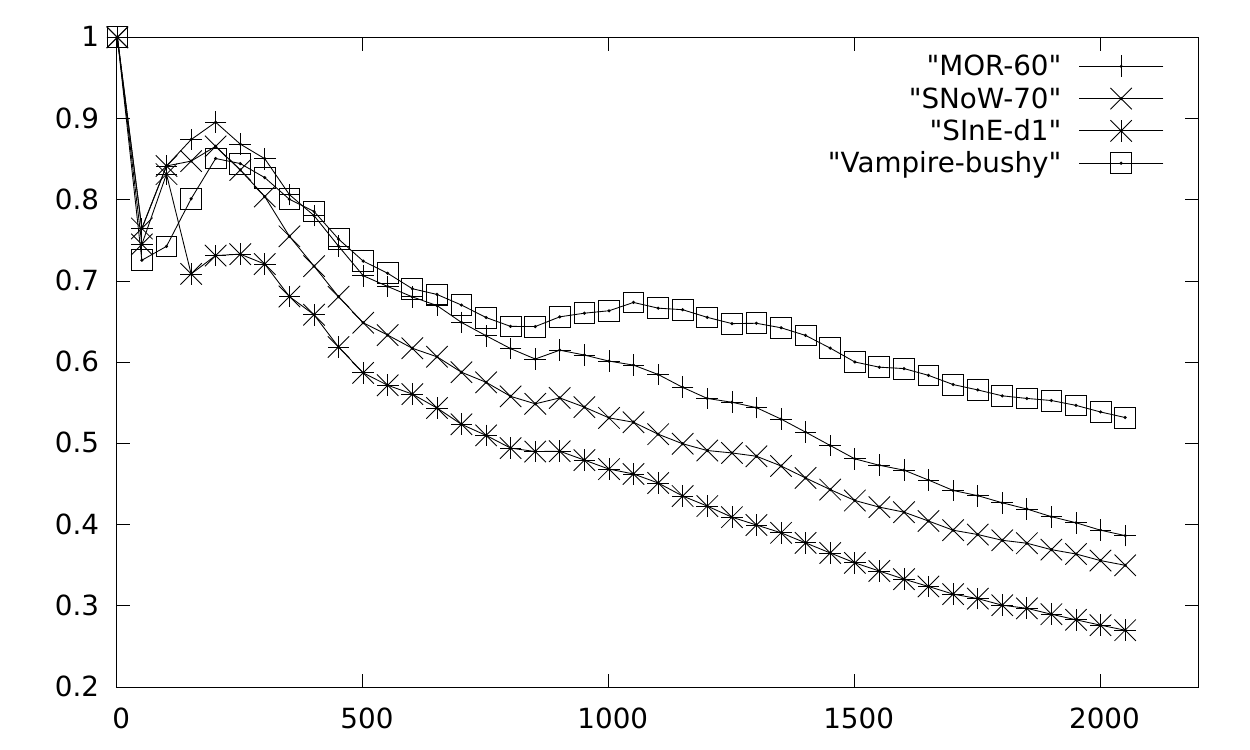}
\end{center}
\caption{Performance of the algorithms on initial segments of MPTP2078.}
\label{Comp1}
\end{figure}

\begin{figure}[!htb]
\begin{center}
\includegraphics[width=12cm]{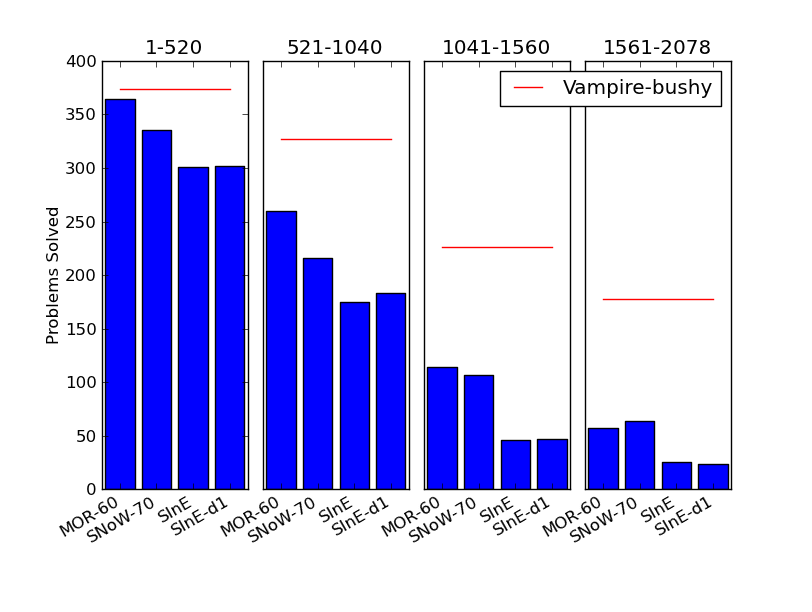}
\end{center}
\caption{Average performance of the algorithms
on four successive equally sized segments of MPTP2078.}
\label{Comp}
\end{figure}

Finally, Figure~\ref{Comp1} and Figure~\ref{Comp} compare the
cumulative and average performance of the algorithms (combined with
ATPs) at different points of the MPTP2078 benchmark, using the
chronological ordering of the MPTP2078 problems.  The average
available number of premises for the theorems ordered chronologically
grows linearly (the earlier theorems and definitions become eligible
premises for the later ones), making the later problems harder on
average.  Figure~\ref{Comp1} shows the performance computed on initial
segments of problems using step value of 50. The last value (for 2050)
corresponds to the performance of the algorithms on the whole MPTP2078
(0.38 for \MOR{}-60), while for example the value for 1000 (0.60 for
\MOR{}-60) shows the performance of the algorithms on the first 1000
MPTP2078 problems.
Figure~\ref{Comp} compares the average performance of the algorithms
when the problems are divided into four successive segments of equal
size. Note that even with the precise use of the \MML{} premises the
problems do not have uniform difficulty across the benchmark, and on
average, even the bushy versions of the later problems get harder. To
visualize this, we also add the values for \Vampire{}-bushy to the
comparison.

Except from small deviations, the ratio of solved problems decreases
for all the algorithms. \Vampire{}/\MOR-60 is able to keep up with
\Vampire{}-bushy in the range of the initial 800 problems, and after
that the human selection increasingly outperforms all the
algorithms. Making this gap as small as possible is an obvious challenge on the path
to strong automated reasoning in general mathematics.

\section{Conclusion and Future Work}
\label{Conclusion}

The performance of automated theorem
proving over real-world mathematics has been significantly improved by using detailed minimized
formally-assisted analysis of a large corpus of theorems and proofs,
and by using improved prediction algorithms.  In particular, it was demonstrated that premise
selection based on learning from exact previous proof dependencies
improves the ATP performance in large mathematical theories by about
44\% when using off-the-shelf learning methods like naive Bayes in
comparison with state-of-the-art general premise-selection heuristics
like SInE.  It was shown that this can be further improved to about 50\% when employing
state-of-the-art kernel-based learning methods.

Automated reasoning in large mathematical libraries is becoming a complex AI
field, allowing interplay of very different AI techniques. Manual
tuning of strategies and heuristics does not scale to large
complicated domains, and data-driven approaches are becoming very useful
in handling such domains. At the same time, existing strong learning methods are
typically developed on imprecise domains, where feedback loops between
prediction and automated verified confirmation as done for example in \MaLARea{} are
not possible. The stronger such AI systems become, the closer we get
to formally assisted mathematics, both in its ``forward'' and
``reverse'' form. And this is obviously another positive feedback loop
that we explore here: the larger the body of formally expressed and
verified ideas, the smarter the AI systems that learn from them.

The work started here can be improved in many possible ways. While
we have achieved 50\% ATP improvement on large problems by better
premise selection resulting in 824 problems proved within 10 seconds, we know
(from~\ref{dep-evaluation}) that with
a better premise selection it is possible to prove at least 1105
problems. Thus, there is still a great opportunity for improved premise
selection algorithms.
Our
dependency analysis can be finer and faster, and combined with ATP and
machine learning systems, can be the basis for a research tool for experimental
formal (reverse) mathematics. An interesting AI problem that is becoming more and more
relevant as the ATP methods for mathematics are getting stronger, is translation
of the (typically resolution-based) ATP proofs into human-understandable~\cite{Davis81,Rudnicki87} formats used by mathematicians.
We believe that  machine learning from large human-proof corpora like \MML{} is likely to be useful for 
this task, in a similar way to how it is useful for finding relevant premises.

The \MOR{} algorithm has a number of parameterizations that we have
fixed for the experiments done here. Further experiments with
different loss functions could yield better results.  One of the most
interesting parameterizations is the right choice of features for the
formal mathematical domain. So far, we have been using only the
symbols and terms occurring in formulas as their feature
characterizations, but other features are possible, and very likely
used by mathematicians. In particular, for ad hoc problem collections
like the TPTP library, where symbols are used inconsistently across
different problems, formula features that abstract from particular
symbols will likely be needed. Also, the output of the learning
algorithms does not have to be limited to the ranking of premises. In
general, all kinds of relevant problem-solving parameterizations can
be learned, and an attractive candidate for such treatment is the
large set of ATP strategies and options parameterizing the proof
search. With such experiments, a large number of alternative ATP proofs 
are likely to be obtained, and an interesting task is to productively learn
from such a combination of alternative (both human and machine) proofs.
Premise selection is only one instance of the ubiquitous 
\emph{proof guidance} problem, and recent prototypes like the \MaLeCoP{}
system~\cite{UrbanVS11} indicate that guidance obtained by machine learning can considerably help also \emph{inside}
automated theorem provers.

Finally, we hope that this work and the performance numbers obtained
will provide a valuable feedback to the CADE competition organizers:
Previous proofs and theory developments in general are an important
part of real-world mathematics and theorem proving. At present,  the LTB division of CASC does not recognize proofs in the way that we are recognizing them here. Organizing
large-theory competitions that separate theorems from their proofs is
like organizing web search competitions that separate web pages from
their link structure~\cite{BrinP98}. We believe that re-introducing a large-theory
competition that does provide both a large number of theorems and a large number of proofs will
cover this important research direction, and most of all, properly
evaluate techniques that significantly improve the ATP end-user
experience.

\section{Acknowledgment}
\label{Ack}
We would like to thank the anonymous JAR referees for a number of
questions, comments, and insights that helped to significantly improve
the final version of this paper.

\bibliography{atpitm}

\begin{thebibliography}{10}
\providecommand{\url}[1]{\texttt{#1}}
\providecommand{\urlprefix}{URL }

\bibitem{alama-dissertation}
Alama, J.: Formal Proofs and Refutations. Ph.D. thesis, Stanford University
  (2009)

\bibitem{AB+11}
Alama, J., Brink, K., Mamane, L., Urban, J.: Large formal wikis: Issues and
  solutions. In: Davenport, J., Farmer, W., Urban, J., Rabe, F. (eds.)
  Intelligent Computer Mathematics, Lecture Notes in Computer Science, vol.
  6824, pp. 133--148. Springer Berlin / Heidelberg (2011)

\bibitem{AlamaKU12}
Alama, J., K{\"u}hlwein, D., Urban, J.: Automated and human proofs in general
  mathematics: An initial comparison. In: Bj{\o}rner, N., Voronkov, A. (eds.)
  LPAR. Lecture Notes in Computer Science, vol. 7180, pp. 37--45. Springer
  (2012)

\bibitem{alama-mamane-urban2011}
Alama, J., Mamane, L., Urban, J.: Dependencies in formal mathematics:
  Applications and extraction for {Coq and Mizar} (2011), arxiv.org pre-print
  abs/1109.3687

\bibitem{aronszajn50reproducing}
Aronszajn, N.: Theory of reproducing kernels. Transactions of the American
  Mathematical Society  68 (1950)

\bibitem{BC04}
Bertot, Y., Cast\'eran, P.: Interactive Theorem Proving and Program
  Development. Coq'Art: The Calculus of Inductive Constructions. Texts in
  Theoretical Computer Science, Springer Verlag (2004)

\bibitem{BlanchetteBN11}
Blanchette, J.C., Bulwahn, L., Nipkow, T.: Automatic proof and disproof in
  {Isabelle/HOL}. In: Tinelli, C., Sofronie-Stokkermans, V. (eds.) FroCos.
  Lecture Notes in Computer Science, vol. 6989, pp. 12--27. Springer (2011)

\bibitem{BrinP98}
Brin, S., Page, L.: The anatomy of a large-scale hypertextual web search
  engine. Computer Networks  30(1-7),  107--117 (1998)

\bibitem{Carlson1999}
Carlson, A., Cumby, C., Rizzolo, N., Rosen, J., Roth, D.: {SNoW user manual}
  (1999), \url{http://l2r.cs.uiuc.edu/~cogcomp/software/snow-userguide.pdf}

\bibitem{Davis81}
Davis, M.: Obvious logical inferences. In: Hayes, P.J. (ed.) IJCAI. pp.
  530--531. William Kaufmann (1981)

\bibitem{mizar-in-a-nutshell}
Grabowski, A., Korni{\l}owicz, A., Naumowicz, A.: Mizar in a nutshell. Journal
  of Formalized Reasoning  3(2),  153--245 (2010)

\bibitem{Harrison96}
Harrison, J.: Optimizing proof search in model elimination. In: McRobbie, M.A.,
  Slaney, J.K. (eds.) CADE. Lecture Notes in Computer Science, vol. 1104, pp.
  313--327. Springer (1996)

\bibitem{HarrisonSA06}
Harrison, J., Slind, K., Arthan, R.: {HOL}. In: Wiedijk, F. (ed.) The Seventeen
  Provers of the World. Lecture Notes in Computer Science, vol. 3600, pp.
  11--19. Springer (2006)

\bibitem{HoderV11}
Hoder, K., Voronkov, A.: Sine qua non for large theory reasoning. In: Bjørner,
  N., Sofronie-Stokkermans, V. (eds.) Automated Deduction – CADE-23, Lecture
  Notes in Computer Science, vol. 6803, pp. 299--314. Springer Berlin /
  Heidelberg (2011)

\bibitem{mizar-first-30}
Matuszewski, R., Rudnicki, P.: {Mizar}: the first 30 years. Mechanized
  Mathematics and Its Applications  4,  3--24 (2005)

\bibitem{MengP08}
Meng, J., Paulson, L.C.: Translating higher-order clauses to first-order
  clauses. J. Autom. Reasoning  40(1),  35--60 (2008)

\bibitem{NipkowPW02}
Nipkow, T., Paulson, L.C., Wenzel, M.: {Isabelle/HOL} - A Proof Assistant for
  Higher-Order Logic, Lecture Notes in Computer Science, vol. 2283. Springer
  (2002)

\bibitem{PaulsonS07}
Paulson, L.C., Susanto, K.W.: Source-level proof reconstruction for interactive
  theorem proving. In: Schneider, K., Brandt, J. (eds.) TPHOLs. Lecture Notes
  in Computer Science, vol. 4732, pp. 232--245. Springer (2007)

\bibitem{PeaseS07}
Pease, A., Sutcliffe, G.: First order reasoning on a large ontology. In:
  Sutcliffe, G., Urban, J., Schulz, S. (eds.) ESARLT. CEUR Workshop
  Proceedings, vol. 257. CEUR-WS.org (2007)

\bibitem{Vampire}
Riazanov, A., Voronkov, A.: The design and implementation of {VAMPIRE}. AI
  Commun.  15(2-3),  91--110 (2002)

\bibitem{MLPprob2}
Richard, M.D., Lippmann, R.P.: {Neural Network Classifiers Estimate Bayesian a
  posteriori Probabilities}. Neural Computation  3(4),  461--483 (2010)

\bibitem{Rifkin2003}
Rifkin, R., Yeo, G., Poggio, T., Rifkin, R., Yeo, G., Poggio, T.: {Regularized
  Least-Squares Classification}. In: Suykens, J., Horvath, G., Basu, S.,
  Micchelli, C., Vandewalle, J. (eds.) Advances in Learning Theory: Methods,
  Model and Applications NATO Science Series III: Computer and Systems
  Sciences, vol. 190, pp. 131--154. IOS Press (2003)

\bibitem{Rudnicki87}
Rudnicki, P.: Obvious inferences. J. Autom. Reasoning  3(4),  383--393 (1987)

\bibitem{Schoelkopf2001}
Schoelkopf, B., Herbrich, R., Williamson, R., Smola, A.J.: {A Generalized
  Representer Theorem}. In: Helmbold, D., Williamson, R. (eds.) Proceedings of
  the 14th Annual Conference on Computational Learning Theory. pp. 416--426.
  Berlin, Germany (2001)

\bibitem{LearningKernels}
Scholkopf, B., Smola, A.J.: Learning with Kernels: Support Vector Machines,
  Regularization, Optimization, and Beyond. MIT Press, Cambridge, MA, USA
  (2001)

\bibitem{Shalev-ShwartzSSC11}
Shalev-Shwartz, S., Singer, Y., Srebro, N., Cotter, A.: {Pegasos: primal
  estimated sub-gradient solver for SVM}. Math. Program.  127(1),  3--30 (2011)

\bibitem{KMB}
Shawe-Taylor, J., Cristianini, N.: Kernel Methods for Pattern Analysis.
  Cambridge University Press, New York, NY, USA (2004)

\bibitem{simpson-sosoa}
Simpson, S.G.: Subsystems of Second Order Arithmetic. Perspectives in
  Mathematical Logic, Springer, 2 edn. (2009)

\bibitem{solovay-fom-email}
Solovay, R.: {AC} and strongly inaccessible cardinals. Available on the
  Foundations of Mathematics archives at
  \url{http://www.cs.nyu.edu/pipermail/fom/2008-March/012783.html} (March 29
  2008)

\bibitem{tarski}
Tarski, A.: On well-ordered subsets of any set. Fundamenta Mathematicae  32,
  176--183 (1939)

\bibitem{PL10}
Tsivtsivadze, E., Pahikkala, T., Boberg, J., Salakoski, T., Heskes, T.:
  Co-regularized least-squares for label ranking. In: H{\"u}llermeier, E.,
  F{\"u}rnkranz, J. (eds.) Chapter in Preference Learning Book). pp. 107--123
  (2010)

\bibitem{Urban06}
Urban, J.: {MPTP} 0.2: Design, implementation, and initial experiments. J.
  Autom. Reasoning  37(1-2),  21--43 (2006)

\bibitem{UrbanHV10}
Urban, J., Hoder, K., Voronkov, A.: Evaluation of automated theorem proving on
  the {Mizar} mathematical library. In: Fukuda, K., van~der Hoeven, J., Joswig,
  M., Takayama, N. (eds.) ICMS. Lecture Notes in Computer Science, vol. 6327,
  pp. 155--166. Springer (2010)

\bibitem{abs-1109-0616}
Urban, J., Rudnicki, P., Sutcliffe, G.: {ATP} and presentation service for
  {Mizar} formalizations. CoRR  abs/1109.0616 (2011)

\bibitem{UrbanS10}
Urban, J., Sutcliffe, G.: Automated reasoning and presentation support for
  formalizing mathematics in {Mizar}. In: Autexier, S., Calmet, J., Delahaye,
  D., Ion, P.D.F., Rideau, L., Rioboo, R., Sexton, A.P. (eds.)
  AISC/MKM/Calculemus. Lecture Notes in Computer Science, vol. 6167, pp.
  132--146. Springer (2010)

\bibitem{US+08}
Urban, J., Sutcliffe, G., Pudl{\'a}k, P., Vyskocil, J.: {MaLARea}
  {SG1}--machine learner for automated reasoning with semantic guidance. In:
  Armando, A., Baumgartner, P., Dowek, G. (eds.) IJCAR. Lecture Notes in
  Computer Science, vol. 5195, pp. 441--456. Springer (2008)

\bibitem{UrbanVS11}
Urban, J., Vyskocil, J., Step{\'a}nek, P.: {MaLeCoP}: Machine learning
  connection prover. In: Br{\"u}nnler, K., Metcalfe, G. (eds.) TABLEAUX.
  Lecture Notes in Computer Science, vol. 6793, pp. 263--277. Springer (2011)

\end{thebibliography}
\bibliographystyle{splncs03}

\end{document}